\documentclass[a4paper,fleqn]{cas-dc}

\usepackage[numbers]{natbib}
\usepackage{enumitem}
\usepackage{diagbox}
\usepackage{xcolor} 
\usepackage{graphicx}
\usepackage[utf8]{inputenc}
\usepackage{tabularx}
\usepackage{amsmath}
\usepackage{subfloat}
\usepackage{underscore}
\usepackage{url}
\usepackage{algorithm}
\usepackage{algorithmicx}
\usepackage{algpseudocode}
\usepackage{soul}
\usepackage{textcomp}
\usepackage{amssymb}
\usepackage{epstopdf}
\usepackage{subfigure}
\begin{document}
\let\WriteBookmarks\relax
\def\floatpagepagefraction{1}
\def\textpagefraction{.001}
\shorttitle{VAE-Bayesian Deep Learning for Solar Generation Forecasting}
\title[mode=title]{A VAE-Bayesian Deep Learning Scheme for Solar Generation Forecasting based on Dimensionality Reduction}
\shortauthors{Devinder Kaur et~al.}
\author[1]{Devinder Kaur}
\ead{devinderkaur@deakin.edu.au}

\address[1]{School of Engineering, Deakin University, Australia}

\author[2]{Shama Naz Islam}
\ead{shama.i@deakin.edu.au}

\address[2]{School of Engineering, Deakin University, Australia}
\author[3]{Md. Apel Mahmud}
\ead{md.a.mahmud@northumbria.ac.uk}

\address[3]{Faculty of Engineering and Environment, Northumbria University, Newcastle Upon Tyne, UK}

\author[4]{Md. Enamul Haque}
\ead{enamul.haque@deakin.edu.au}

\address[4]{School of Engineering, Deakin University, Australia}
\author[5]{Adnan Anwar}
\ead{adnan.anwar@deakin.edu.au}

\address[5]{School of Information Technology, Deakin University, Geelong 3216, Strategic Centre for Cyber Security Research Institute (CSRI), Australia}

\begin{abstract}
	The advancement of distributed generation technologies in modern power systems has led to a widespread integration of renewable power generation at customer side. 
	However, the intermittent nature of renewable energy poses new challenges to the network operational planning with underlying uncertainties. 
	This paper proposes a novel Bayesian probabilistic technique for forecasting renewable solar generation by addressing data and model uncertainties by integrating bidirectional long short-term memory (BiLSTM) neural networks while compressing the weight parameters using variational autoencoder (VAE).
	Existing Bayesian deep learning methods suffer from high computational complexities as they require to draw a large number of samples from weight parameters expressed in the form of probability distributions. The proposed method can deal with uncertainty present in model and data in a more computationally efficient manner by reducing the dimensionality of model parameters. The proposed method is evaluated using quantile loss, reconstruction error, and deterministic forecasting evaluation metrics such as root-mean square error.
	It is inferred from the numerical results that VAE-Bayesian BiLSTM outperforms other probabilistic and deterministic deep learning methods for solar power forecasting in terms of accuracy and computational efficiency for different sizes of the dataset. 
\end{abstract}
\begin{highlights}
\item A Bayesian BiLSTM model for solar generation forecasting is developed by implementing mean field VI to obtain a tractable approximation of the posterior distribution of weight parameters. 
	\item To improve the computational efficiency of the Bayesian BiLSTM forecasting, a VAE-based dimensionality reduction technique is proposed to compress the sample size of weight parameters using variational encoding layer. The information loss is monitored using reconstructed data from the decoding layer.
	\item The proposed VAE-Bayseian BiLSTM model is evaluated with variable data sizes for solar generation and compared with a number of probabilistic deep learning methods. The proposed method demonstrates superior results in terms of pinball losses, prediction intervals, and root mean square error (RMSE), while achieving a much lower reconstruction loss for weight parameters.
\end{highlights}
\maketitle
\section{Introduction}
The advancement in distributed
generation (DG) technologies has led to a widespread integration of distributed renewable energy sources (RES) such as solar panels and wind turbines at the user end \cite{el2004distributed}. 
However, the intermittent RES generation
often pose new challenges to distribution networks due to external factors such as weather changes and user behavior \cite{sun2019using,gandhi:2020}.
In this context, energy forecasting has a crucial role to play in planning and managing the operations of modern power grid often regarded as smart grid (SG) systems \cite{kaur2020energy}. 
Since the RES is an intermittent source of energy generation, its accurate forecasting ahead of time can be of key importance to increase the operational efficiency of the grid and make it more sustainable \cite{ALKHAYAT2021100060}. 

Recent research on energy forecasting highly focuses on deep learning-based methods such as recurrent neural networks (RNN) and long short-term memory (LSTM)  which utilize the available big data from SG infrastructure \cite{kong2017short,kaur2019smart}. 
Furthermore, bidirectional RNN combined with LSTM formulates the architecture for bidirectional LSTM (BiLSTM) and exhibits the advantage of long-term memory for tackling vanishing gradients in addition to an extensive learning process. Bidirectional processing leads to proper exploitation of all data points and thus, improves the forecasting accuracy. In this regard, BiLSTM-based techniques have been implemented in \cite{jahangir2020deep}, \cite{toubeau2018deep} to improve the learning process for the application of short-term (e.g., few days ahead) energy forecasting in power markets. Deep learning methods are often incorporated with optimisation techniques to improve the effectiveness of the training process \cite{amjady:2011}. Although the extensive learning often leads to increased computation time, a trade-off needs to be maintained between the forecasting accuracy and time complexity of the model. In addition, standard deep learning forecasting methods such as RNN and its variants generate point forecasts and often struggle with challenges of overfitting and appropriate hyper-parameter tuning.
Various regularization techniques have been proposed in the past to tackle the issue of overfitting, such as dropout and early stopping \cite{srivastava2014dropout}, \cite{gal2016dropout}.

Despite the recent advancements in deep learning techniques, the uncertainties in the energy data due to external factors such as changes in weather, as well as the uncertainty presented in the model parameters pose a significant challenge to accurate energy forecasting. In this context, probabilistic approaches have been widely incorporated in energy forecasting due to their ability to manage uncertainties. For example, the authors in \cite{Bessa:2015} utilized probabilistic methods integrated with vector autoregression for solar power forecasting. 
In a similar manner, non-parametric Bayesian approach for renewable energy forecasting is implemented in \cite{bracale2013bayesian, xie2018nonparametric,ning2019data}. An extreme learning machine-based probabilistic approach has been adopted in \cite{Wan:2014} for wind power forecasting. On the other hand, the authors in \cite{ZHU2022100199} used an ensemble decomposition approach to separate wind time series data and then forecasted using support vector regression model. The authors in \cite{Thomas:2020} adopted analog ensemble-based probabilistic forecasting for solar generation. Due to their ability to address uncertainty, probabilistic methods can be integrated with deep neural layers to exploit the advantages of both the probabilistic and deep learning techniques \cite{gal2016uncertainty}. Kendall \textit{et. al.} in \cite{kendall2017uncertainties} highlighted the problem of uncertainty and proposed Bayesian neural networks (BNN) based on the conditional probability to quantify the epistemic or model uncertainty, represented by a probability distribution on weight parameters of neural network layers. Though the epistemic uncertainty can be explained away by adding more data, uncertainty inherited from the input data in the form of noise or aleatoric uncertainty cannot be reduced with more data.

Very recently, Bayesian probabilistic approach integrated with deep learning methods have been proposed for demand forecasting \cite{yang2019bayesian}. Such methods generate forecasts in the form
of prediction intervals (PIs) in which the forecast is predicted to lie with a certain probability. PIs are used to quantify the uncertainties associated with model parameters and energy consumption in probabilistic deep learning, contrary to the traditional deep neural networks that are deterministic in nature \cite{sun2019using}. Bayesian neural networks (BNNs) have been proposed in \cite{yacef:2012} for solar irradiation forecasting. The authors in \cite{raza:2018} apply Bayesian model averaging to estimate the predictors of ensemble forecasting of solar power generation. BNNs have been investigated in \cite{yang2019bayesian} for load forecasting applications and it has been observed that BNNs take more time to converge in comparison to other probabilistic deep learning methods due to large dimensions resulting from probability distribution of model parameters.

Furthermore, variational inference (VI) techniques have emerged as an effective method to obtain posterior distribution of weight parameters by minimizing the divergence between true and approximate posterior distributions in Bayesian probabilistic methods \cite{zhang2018advances}. The authors in \cite{liu:2019} implemented VI with convolutional gate recurrent unit (GRU) network for forecasting solar irradiation. VI is integrated with LSTM in \cite{zhang:2020} for solar irradiation forecasting while ensuring that training data is not shared among multiple solar generation sites. Though VI can offer a tractable solution to obtain posterior distributions of model parameters, the use of probability distributions significantly expands the model parameter
space, especially for weight matrices. Thus, the ability of Bayesian deep learning techniques to deal with model and stochastic uncertainties comes at the cost of computational complexity arising from sample size of weight parameters.

In this context, variational autoencoders (VAE), widely used in the existing deep learning literature for dimensionality reduction \cite{Pramod:2019}, can be implemented to improve the computational efficiency of Bayesian deep learning techniques. VAE represents an encoder with probability distributions which transforms the input data to adopt a lower dimensional structure \cite{kingma2013auto}. VAEs integrated with deep learning have been widely utilized in different power system applications such as fault and anomaly detection in time series energy data \cite{pereira2018unsupervised,wang2020variational,biswas2020devlearn}. Recently, VAE is implemented in \cite{Dairi:2020} to forecast the solar generation data. VAE is also used as a dimensionality reduction technique to manage the large dimensions of the input layer when multiple time lags are considered and it has been shown to improve the efficiency of forecasting methods \cite{kaur2021ICC}.
\subsection{Motivation and research contributions}
Existing research has focused on the benefits of Bayesian deep learning methods to address the issue of uncertainty. However, the integration of advanced deep learning
methods such as BiLSTM and VAE with Bayesian models while representing such models with more tractable approximations
of weight parameters remain an open problem for energy forecasting applications. In addition, an effective solution to manage the increasing number of weight parameters for Bayesian deep learning methods have not been proposed yet. 

To address the aforementioned challenges, we propose a new Bayesian BiLSTM forecasting technique integrated with VAE to tackle the uncertainties in the granular renewable energy generation data along with high dimensionality in the model parameters of probabilistic methods. The specific contributions are as follows:
\begin{enumerate}
	\item A Bayesian BiLSTM model for solar generation forecasting is developed by implementing mean field VI to obtain a tractable approximation of the posterior distribution of weight parameters. 
	\item To improve the computational efficiency of the Bayesian BiLSTM forecasting, a VAE-based dimensionality reduction technique is proposed to compress the sample size of weight parameters using variational encoding layer. The information loss is monitored using reconstructed data from the decoding layer.
	\item The proposed VAE-Bayseian BiLSTM model is evaluated with variable data sizes for solar generation and compared with a number of probabilistic deep learning methods. The proposed method demonstrates superior results in terms of pinball losses, prediction intervals, and root mean square error (RMSE), while achieving a much lower reconstruction loss for weight parameters.
\end{enumerate}
\subsection{Paper organization}
The rest of the paper is organized in the following manner. Section \ref{sec:problem} outlines the problem formulation regarding weight compression and optimized distribution of weight parameters for forecasting. Section \ref{sec:proposed} elaborates how the proposed VAE-BiLSTM model is implemented for forecasting renewable generation. Numerical evaluation and comparison results for the proposed scheme have been presented in Section \ref{sec:results}. Finally, the paper is concluded in Section \ref{sec:conclusions}.

\begin{figure}[!t]
	\centering
	\includegraphics[height=0.185\textheight,width=0.45\textwidth]{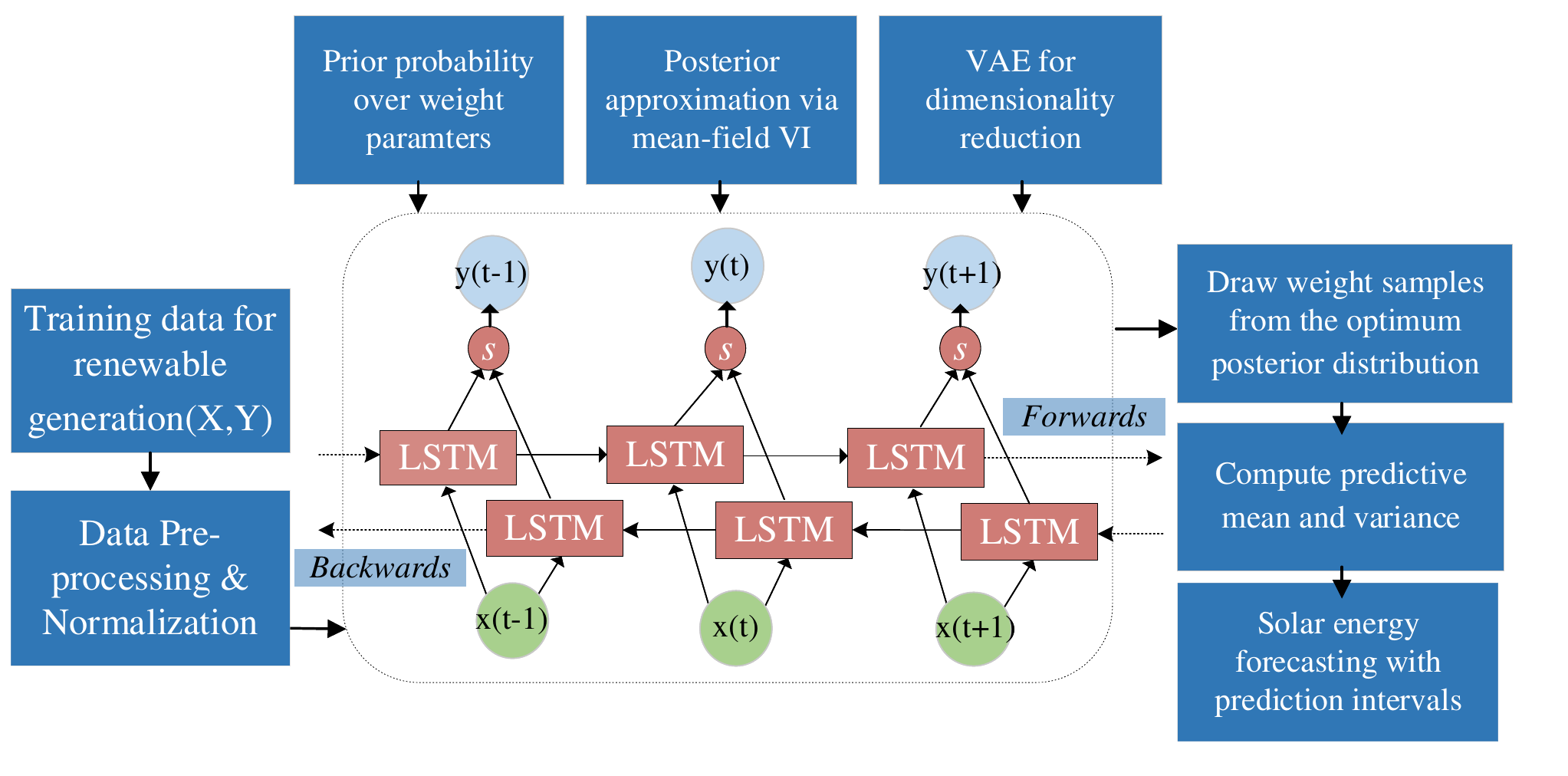}
	\caption{System model of the proposed VAE-Bayesian BiLSTM scheme}
	\label{flow}
\end{figure}
\begin{figure}[!h]
	\begin{center}
	\includegraphics[height=0.185\textheight,width=0.45\textwidth]{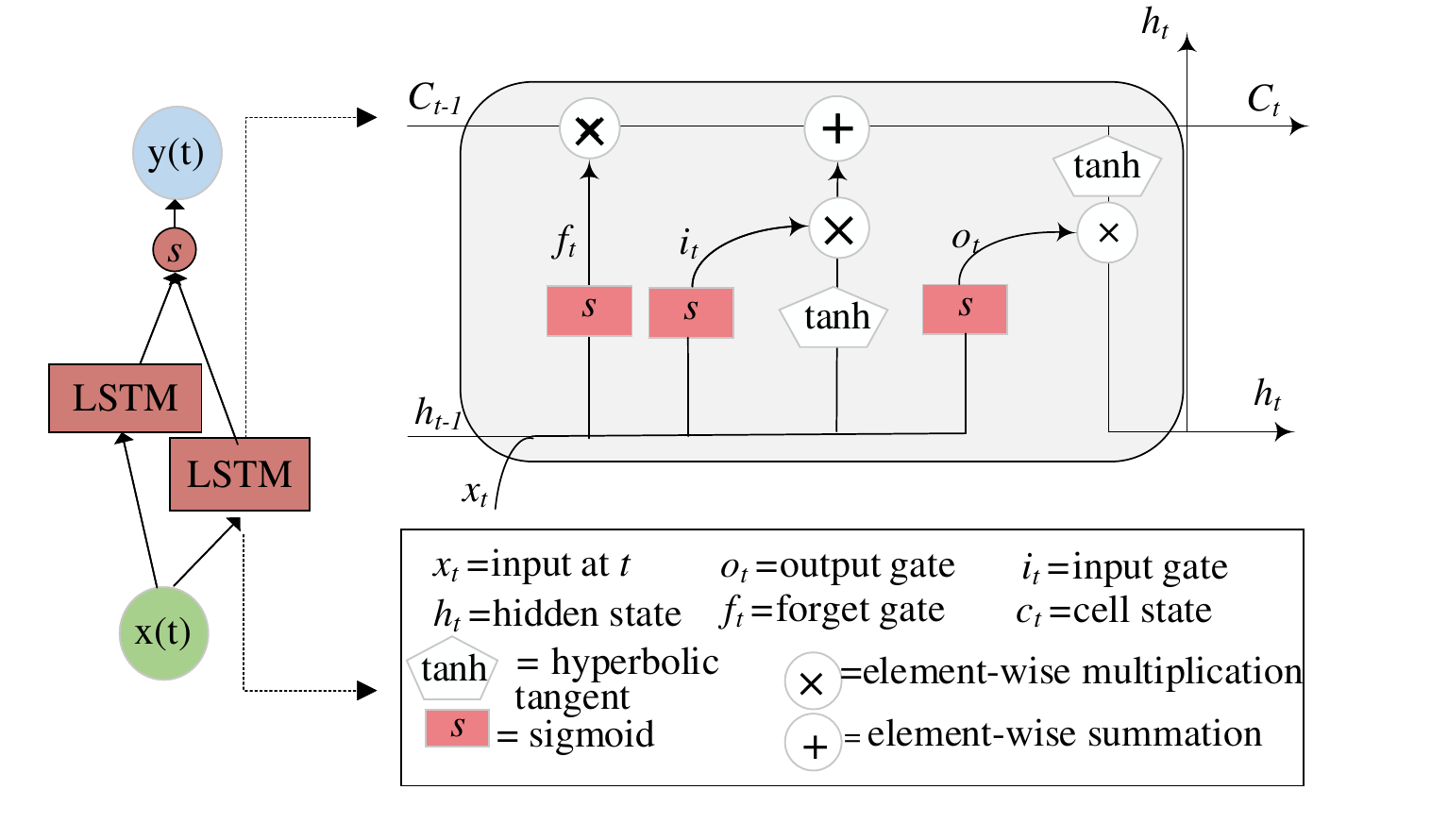}
	\caption{BiLSTM architecture}
	\label{blstm}
	\end{center}
\end{figure}
\begin{figure}[!t]
	\begin{center}
	\includegraphics[width=0.45\textwidth]{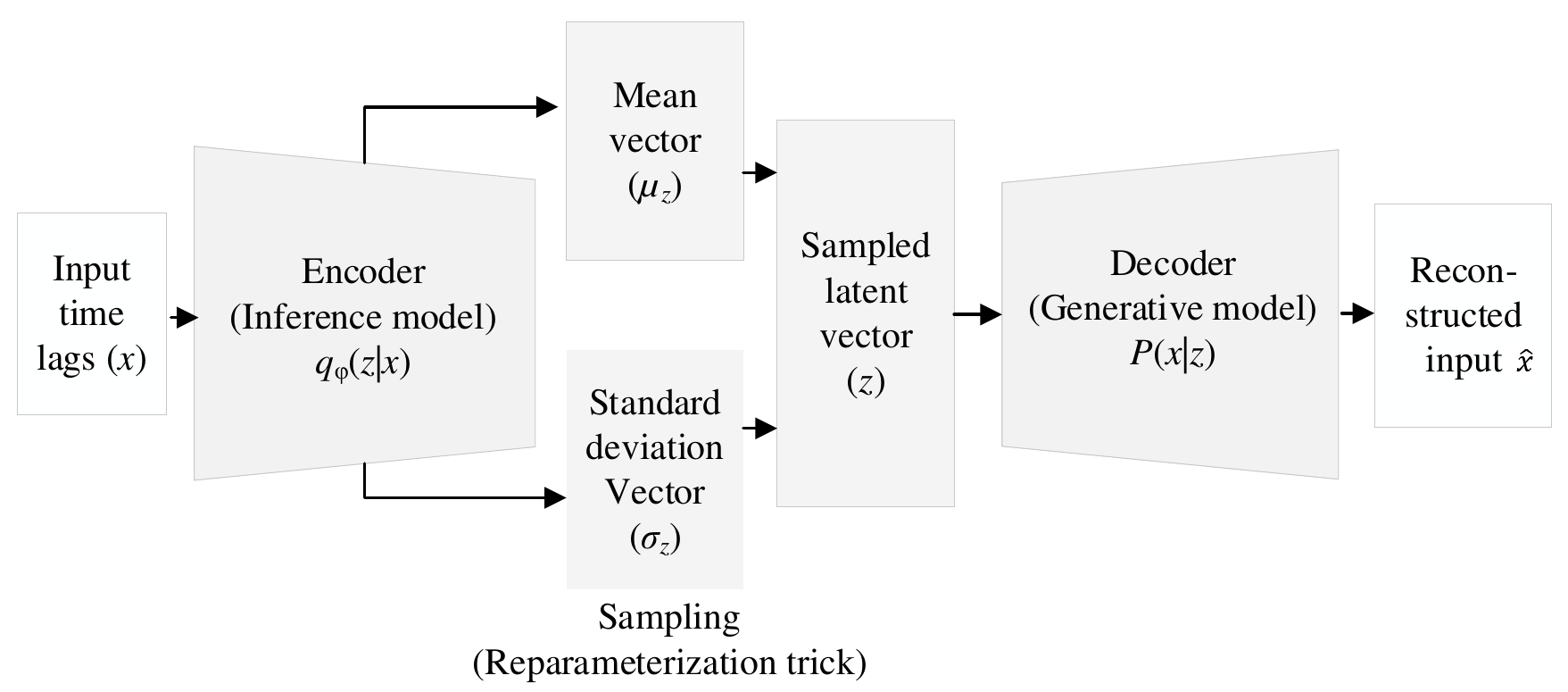}
	\caption{Overview of variational autoencoder}
	\label{vae}
\end{center}
\end{figure}
\section{Problem Formulation}\label{sec:problem}
For a given input data $D$ = ($x_t, y_t$), where $x_t$ denotes historical renewable generation data with time lags, $y_t$ denotes true observations of renewable generation at interval $t$, and $p(w)$ is the prior distribution on weight parameters $w$ of the BiLSTM network, Bayesian probabilistic model is represented using following equation \cite{zhang2018advances}:
\begin{equation}\label{eq:bayes}
p(w|D)=\frac{p(D|w)p(w)}{p(D)}
\end{equation}
where $p(w|D)$ represents the posterior probability of weight parameters given observed data using Bayes rule of conditional probability and $p(D|w)$ denotes the likelihood of observed data given weight parameters. Furthermore, $p(D)$ represents the marginal probability for the observed parameters. Therefore, equation (\ref{eq:bayes}) is expressed as \cite{zhang2018advances}:
\begin{equation}\label{3eq:posterior}
p(w|D)=\frac{p(D|w)p(w)}{\int(p(D,w)p(w)dw}
\end{equation}
\indent where $p(D,w)$ is the joint probability distribution of $D$ and $w$. Due to the large number of data values presented in the integral of (\ref{3eq:posterior}), it becomes intractable to analytically compute the exact posterior.
So, instead of computing the true posterior, an approximated distribution $q(w,\theta)$ parameterized over $\theta$ is computed using mean field VI(MFVI). Here $\theta$ represents the set of mean and standard deviation of the Gaussian distribution for the approximate posterior.

In MFVI, a factorized distribution is used as the approximate posterior distribution. Thus, the approximate posterior probability distribution satisfies the following condition \cite{zhang2018advances}:

\begin{equation}\label{mfvi}
q_{\theta}(w|D) = {\prod_{t=1}^{N}}q_{\theta}(w_{t}|D_{t})
\end{equation}
where $q_{\theta}(w_{t}|D_{t})$ represents the factorized posterior distribution for weight parameters $w$ given observed data $D$ over $N$ samples.
Then, the difference between real and approximated posterior is minimized using  Kullback-Leibler (KL) divergence \cite{kendall2017uncertainties}, given as:
\begin{equation}\label{eq:KL}
KL(q_\theta(w)||p(w))=  \log{p(D)}- \mathop{\mathbb{E}}_{q_\theta(w)}\log \frac{p(D,w)}{q_\theta(w)}
\end{equation}
\begin{equation}
\textrm{s.t.} \quad 
KL(q_\theta(w)||p(w)) \ge 0 \\
\end{equation}

KL divergence  is asymmetric in nature and also referred as information gain or relative entropy.
Finally, the cost function for the proposed scheme is to minimize the KL divergence as in:
\begin{equation}\label{eq:optKL}
arg \ min_{q_\theta(w)} KL(q_\theta(w)||p(w))
\end{equation}

To minimize KL divergence, evidence lower bound (ELBO) is maximized. ELBO is defined as lower bound on the log marginal probability of observed data points \cite{kendall2017uncertainties}. The marginal probability of data points can be expressed as following:
\begin{equation}
\log p(D) = \log \int p(D,w) dw
\end{equation}
\begin{equation}
\log p(D) = \log  \int \frac{p(D,w) *  q_\theta(w)}{q_\theta(w)} dw
\end{equation}
\begin{equation}
\log p(D) = \log \mathop{\mathbb{E}}_{q_\theta(w)} [\frac{p(D,w)}{q_\theta(w)} ]
\end{equation}
\begin{equation}\label{eq:ELBO}
\log p(D) \ge  \mathop{\mathbb{E}}_{q_\theta(w)} \log \left[\ \frac{p(D,w)}{q_\theta(w)} \right]\
\end{equation}

\noindent where \eqref{eq:ELBO} follows from Jensen's inequality. Comparing \eqref{eq:KL} and \eqref{eq:ELBO}, it is evident that
\begin{equation}\label{elbo}
\log p(D) = ELBO (\theta) + KL(q_\theta(w)||p(w)) 
\end{equation}

Thus, maximizing variational ELBO leads to KL minimization and the optimization problem in \eqref{eq:optKL} is reformulated as:
\begin{equation}
arg \max_{q_{\theta}(w)} ELBO
\end{equation}

Now, the approximated posterior in the aforementioned Bayesian BiLSTM method is a distribution $q_\theta(w)$ rather than a single weight sample. This results into large dimensionality of weight parameters drawn from the optimal distribution. To overcome this issue, VAE is integrated to encode the large sample space during posterior approximation into lower dimensional space representation $z$  using following equation:
\begin{equation}\label{eq:vae}
p(z|x)=\frac{p(x|z)p(z)}{p(x)}
\end{equation}
where $p(z|x)$ represents the encoder on the input samples $x$ (weight parameters from Bayesian BiLSTM), $p(z)$ is the marginal distribution of the encoded data and $p(x|z)$ denotes the decoder to represent the likelihood of reconstructed data over weight parameters in a VAE process.
$p(x)$ symbolizes the marginalized probability of input data. Thus \eqref{eq:vae} can further be defined as:
\begin{equation}\label{eq:vae}
p(z|x)=\frac{p(x|z)p(z)}{\int(p(x,z)p(z)dz}
\end{equation}
\noindent where $p(x,z)$ is the joint distribution of input and encoded data. VI can be used to approximate posterior distribution $q_{\phi}(z|x)$ over parameters $\phi$ as a representative of the true posterior distribution $p(z|x)$. Here $\phi$ denotes the set of mean and standard deviation of posterior distribution of the encoder.
In this context, the cost function for VAE consists of reconstruction loss and Variational loss. The reconstruction loss is due to error in reconstructing the actual data from the decoding process. On the other hand, the variational loss of VAE is computed using KL divergence. Thus, the aggregated loss for VAE is defined as:
\begin{equation}\label{recon}
\tau_{loss} = \mathop{\mathbb{E}}_{q_\phi(z)}[\log p(x|z)]+KL(q_\phi(z|x)||p(z))
\end{equation}
\noindent where the first term denotes the likelihood of reconstructed data at the decoder and the second term is the KL divergence between the approximated and real posterior distribution of the latent variable, given by:
\begin{equation}\label{eq:KLphi}
KL(q_\phi(z|x)||p(z))=-\int q_\phi(z|x)\log\left[\frac{p(z)}{q_\phi(z|x)}\right]dz
\end{equation}
So, the optimised encoder can be obtained by solving:
\begin{equation}
\arg \min_{q_\phi(z)} \tau_{loss}
\end{equation}

The aforementioned system model illustrating the integration of VAE with Bayesian BiLSTM is presented in Fig. \ref{flow}. The
architecture of BiLSTM network and VAE are shown in Fig. \ref{blstm} and \ref{vae}, respectively.

\section{Proposed scheme}\label{sec:proposed}
\begin{figure}[!t]
	\centering
	\includegraphics[width=0.46\textwidth]{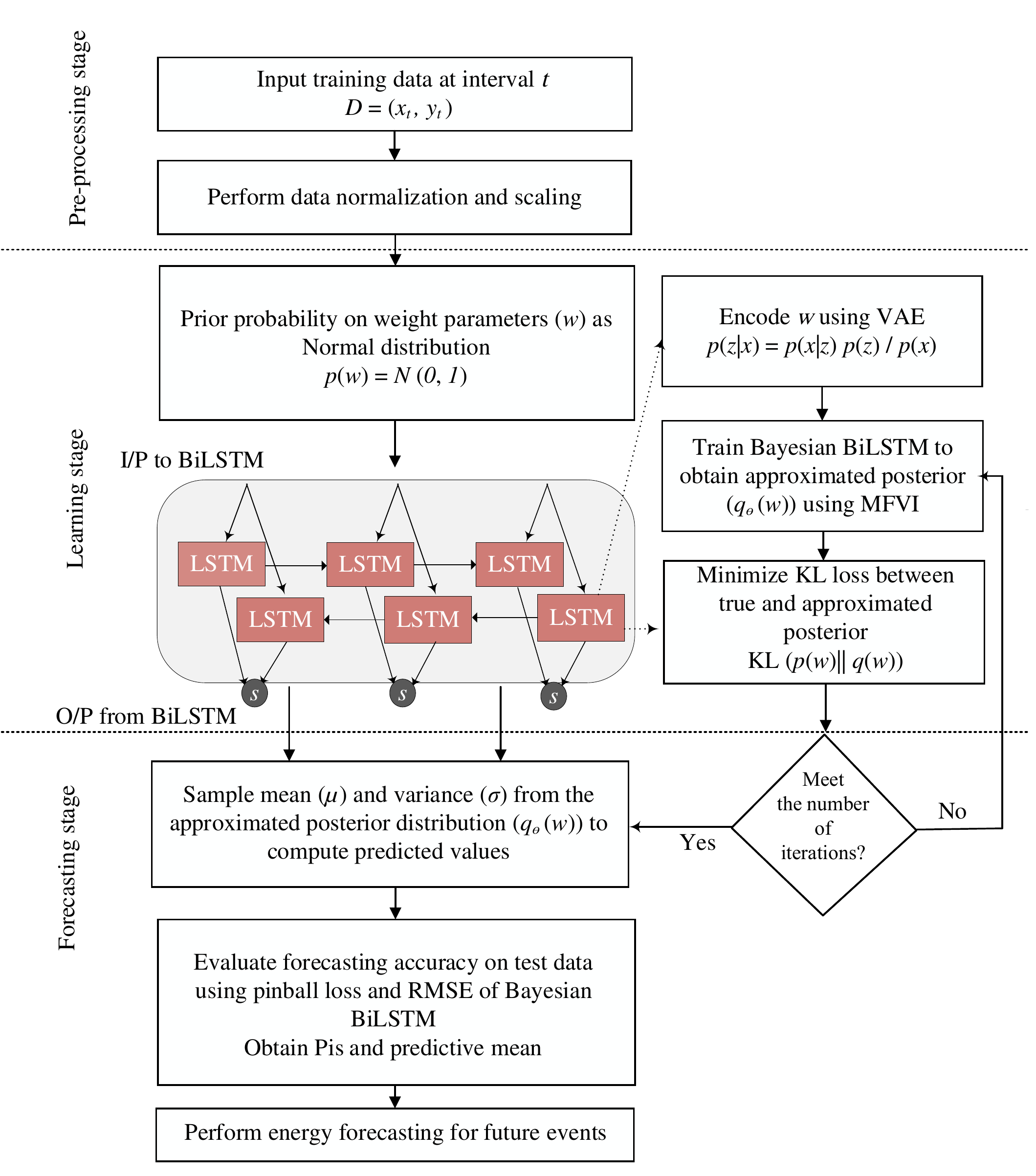}
	\caption{Flow chart of the proposed VAE-Bayesian BiLSTM algorithm}
	\label{flowchart}
\end{figure}

This section outlines the proposed framework for the integrated VAE-Bayesian BiLSTM probabilistic forecasting method. 
The methodology for the proposed algorithm is illustrated in Fig. \ref{flowchart} in the form of a detailed flow chart.
In this paper, Bayesian BiLSTM method is employed to address data and
model uncertainties while VAE addresses high dimensionality of weight parameters due to Bayesian probabilistic method implemented on BiLSTM layers.

\begin{algorithm}
	\textbf{Input:} Actual observed data, $D=(x_{t},y_{t})$
	\\
	\textbf{Output:} Prediction intervals (PIs), predictive mean ($\hat{y}$)
	\begin{algorithmic}[1]
		\State Obtain observed data $D$;
		\State Scale and normalize $D$ in (-1,1);
		\State Split training data $D_i$ from $D$;
		\State Feed $D_i$ to VAE-Bayesian BiLSTM;
		\Statex\textbf{Encoding-decoding stage using VAE}
		\State Compute prior probability on latent space $p(z)$ using \eqref{eq:Gaussianprior};
		\State Define approximate posterior $q(z|D)$ using \eqref{eq:Gaussianapproxpost};
		\State Optimize posterior distribution of the encoded data at the encoder;
		\State Optimize likelihood of the reconstructed data at the decoder;
		\State Minimize $\tau_{loss}$ using \eqref{eq:reconGaussian} ;
		\Statex	\textbf{Forecasting stage using Bayesian BiLSTM stage}
		\State Sample from z;
		\State \textbf{for (i=1, i $\le$ n, i++) do}
		\State  \indent Formulate prior trainable on weight parameters p(w);
		\State  \indent Use BiLSTM layer as hidden layer;
		\State  \indent Use DenseVariational layer as output layer;
		\State  \indent Approximate independent $p(w|D)$ using \eqref{mfvi};	
		\State  \indent Maximize ELBO using \eqref{elbo};
		\State \textbf{ end for} 
		\State  Compute $\mu$ and $\sigma$ and  obtain PIs;
		\State  Compute predictive mean;
		\State  Evaluate error performance of the forecasting technique;
		\State Evaluate the model performance using time complexity and reduction in $w$;
	\end{algorithmic}
	\caption{VAE-Bayesian BiLSTM framework for probabilistic forecasting}
\end{algorithm}
\subsection{Bayesian BiLSTM}
Bayesian probability incorporated with neural networks gives forecasting results in the form
of prediction intervals to quantify uncertainties associated with model parameters and renewable energy
generation.
Epistemic uncertainty also known as model uncertainty is the uncertainty related to the
model parameters such as weights and biases in a neural network.
Furthermore, 
aleatoric uncertainty is defined as the uncertainty related to data and also referred to as stochastic uncertainty. It captures the noise inherent in the observed data, either due to exogenous factors such as weather and human behavior patterns or from sensor devices during data acquisition. 
Although, it is not feasible to reduce the aleatoric uncertainty with the help of more data, it can be quantified using Bayesian deep learning approach by representing output layer with probability distributions. Furthermore, by optimizing weights and reducing weight uncertainty, prediction accuracy is improved for the future forecasts.

Using proposed Bayesian approach, aleatoric uncertainty is quantified by
placing a prior normal distribution over the objective function. Whereas, epistemic uncertainty
is handled by placing a prior over weight parameters of the BiLSTM network. Furthermore,
the uncertainty can be reduced with the help of hyperparameter tuning and optimizing
the learning process. 

To be specific, both the uncertainties are represented by placing prior on 
weights and objective function using a DenseVariational layer incorporated with BiLSTM hidden layer. Then, the approximate posterior distribution is computed using
MFVI based on evidence provided by the observed data by minimizing the KL loss using \eqref{eq:KL}. In addition, BiLSTM works in forward
and backward passes as illustrated in Fig. \ref{flow}. 
\subsection{VAE for weight compression}
To reduce the sample space for weight parameters, VAE is integrated. VAEs are defined as the class of generative models.
Since VAE tries to learn a distribution from the latent space, $\epsilon$ in the following equation is used to parametrize the VAE network \cite{kingma2013auto} with normal distribution having $\mu_{z}$ as mean and $\sigma_{z}$ as the standard deviation for the latent distribution:
\begin{equation}
z=\mu_{z} + \sigma_{z} \odot \epsilon
\end{equation}

\noindent where $\odot$ represents element-wise multiplication. Based on the assumption of normal distribution, the prior probability distribution of the latent variable can be written as:
\begin{equation}\label{eq:Gaussianprior}
p(z)=\frac{1}{\sqrt{2\pi\sigma_z^2\epsilon^2}}e^{-\frac{(x-\mu_z)^2}{2\sigma_z^2\epsilon^2}}
\end{equation}
Similarly, the probability distribution of the approximated posterior is represented as:
\begin{equation}\label{eq:Gaussianapproxpost}
q_\phi(z|x)=\frac{1}{\sqrt{2\pi\sigma_q^2}}e^{-\frac{(x-\mu_q)^2}{2\sigma_q^2}}
\end{equation}
\noindent where $\mu_q$ and $\sigma_q$ are the mean and standard deviation of the approximated posterior. Substituting the probability distributions of prior and approximated posterior from \eqref{eq:Gaussianprior} and \eqref{eq:Gaussianapproxpost} into \eqref{eq:KLphi}, the KL divergence can be expressed as:
\begin{align}\label{eq:GaussianKL}
KL(q_\phi(z|x)||p(z))&=\int \frac{1}{\sqrt{2\pi\sigma_q^2}}e^{-\frac{(x-\mu_q)^2}{2\sigma_q^2}}\times\nonumber\\
&\log\left[\sqrt{\frac{\sigma_q^2}{\sigma_z^2\epsilon^2}}e^{-\frac{(x-\mu_q)^2}{2\sigma_q^2}-\frac{(x-\mu_z)^2}{2\sigma_z^2\epsilon^2}}\right]dz
\end{align}

After some mathematical manipulations, the aforementioned expression can be simplified as:
\begin{align}\label{eq:GaussianKL}
KL(q_\phi(z|x)||p(z))&=\int \frac{1}{\sqrt{2\pi\sigma_q^2}}e^{-\frac{(x-\mu_q)^2}{2\sigma_q^2}}\left(\log\left[\frac{\sigma_q}{\sigma_z\epsilon}\right]+\right.\nonumber\\
&\left.\frac{(x-\mu_q)^2}{2\sigma_q^2}-\frac{(x-\mu_z)^2}{2\sigma_z^2\epsilon^2}\right) dz
\end{align}

Replacing the integral with expectation, the above equation can be further simplified to:
\begin{align}\label{eq:KLexpect}
KL(q_\phi(z|x)||p(z))&=\mathop{\mathbb{E}}_{q_\phi(z)}\left[\log\left[\frac{\sigma_q}{\sigma_z\epsilon}\right]\right.+\nonumber\\
&\left.\frac{(x-\mu_q)^2}{2\sigma_q^2}-\frac{(x-\mu_z)^2}{2\sigma_z^2\epsilon^2}\right]
\end{align}

Since the term inside logarithm is a constant, \eqref{eq:KLexpect} can be written as:
\begin{align}\label{eq:KLexpect}
KL(q_\phi(z|x)||p(z))&=\log\left[\frac{\sigma_q}{\sigma_z\epsilon}\right]+\frac{1}{2\sigma_q^2}\mathop{\mathbb{E}}_{q_\phi(z)}\left[(x-\mu_q)^2\right]-\nonumber\\
&\frac{1}{2\sigma_z^2\epsilon^2}\mathop{\mathbb{E}}_{q_\phi(z)}\left[(x-\mu_z)^2\right]
\end{align}

Given that the term $\mathop{\mathbb{E}}_{q_\phi(z)}\left[(x-\mu_q)^2\right]$ in \eqref{eq:KLexpect} represents the variance $\sigma_q^2$ of the approximated posterior, the above equation can be simplified to:
\begin{equation}\label{eq:KLGaussianFinal}
KL(q_\phi(z|x)||p(z))= \log\left[\frac{\sigma_q}{\sigma_z\epsilon}\right]+\frac{1}{2}-\frac{1}{2\sigma_z^2\epsilon^2}\mathop{\mathbb{E}}_{q_\phi(z)}\left[(x-\mu_z)^2\right]   
\end{equation}

Now the likelihood probability of the data given the latent variable is represented as:
\begin{equation}\label{eq:likelihood}
p(x|z)=\frac{1}{\sqrt{2\pi\sigma_z^2\epsilon^2}}e^{-\frac{(x-\hat{x})^2}{2\sigma_z^2\epsilon^2}}
\end{equation}

Referring back to \eqref{recon}, the reconstruction loss is written as:
\begin{equation}\label{eq:reconGaussian}
\mathop{\mathbb{E}}_{q_\phi(z)}[\log p(x|z)]\propto  \mathop{\mathbb{E}}_{q_\phi(z)}[-(x-\hat{x})^2]
\end{equation}
\noindent where, $\hat{x}$ represents the reconstructed data samples from the latent distribution. Thus, the aggregated loss function for VAE becomes:
\begin{equation}\label{eq:lossGaussian}
\tau_{loss}= \mathop{\mathbb{E}}_{q_\phi(z)}[-(x-\hat{x})^2]+\log\left[\frac{\sigma_q}{\sigma_z\epsilon}\right]+\frac{1}{2}-\frac{1}{2\sigma_z^2\epsilon^2}\mathop{\mathbb{E}}_{q_\phi(z)}\left[(x-\mu_z)^2\right]
\end{equation}

Finally, the VAE-Bayesian deep learning model is trained using the aforementioned loss function and probabilistic forecasts of the renewable energy generation is obtained. The predictive mean and standard deviation of the predicted distribution is then computed.

Algorithm I represents the pseudo-code for the proposed scheme. 
After performing data pre-processing on input data, it is fed to the VAE stage to encode higher dimensional data into lower dimensions. Data is encoded using variational autoendoer and decoded from latent space to monitor the information loss during the encoding process (lines 5-9).
After achieving the least value of $\tau_{loss}$, the encoded representation is used for forecasting stage employed based on Bayesian probability.
The learning process is mainly carried out with the help of DenseVariational layer integrated with BiLSTM hidden layer (lines 12-15). With the concept of MFVI, posterior distribution is obtained for the compressed weight parameters. The training in forecasting stage is optimized by maximizing the ELBO (line 12). After achieving the maximum ELBO, predictive mean is calculated from the sampled mean and variance and the proposed method is tested based on the evaluation metrics (lines 19-22).

\section{Results and discussions}\label{sec:results}
This section elaborates data description, evaluation criteria, implementation results  and a comparative case study for the proposed VAE-Bayesian BiLSTM method. 
\subsection{Data Description}
\begin{figure}[!b]
	\centering
	\includegraphics[width=0.5\textwidth]{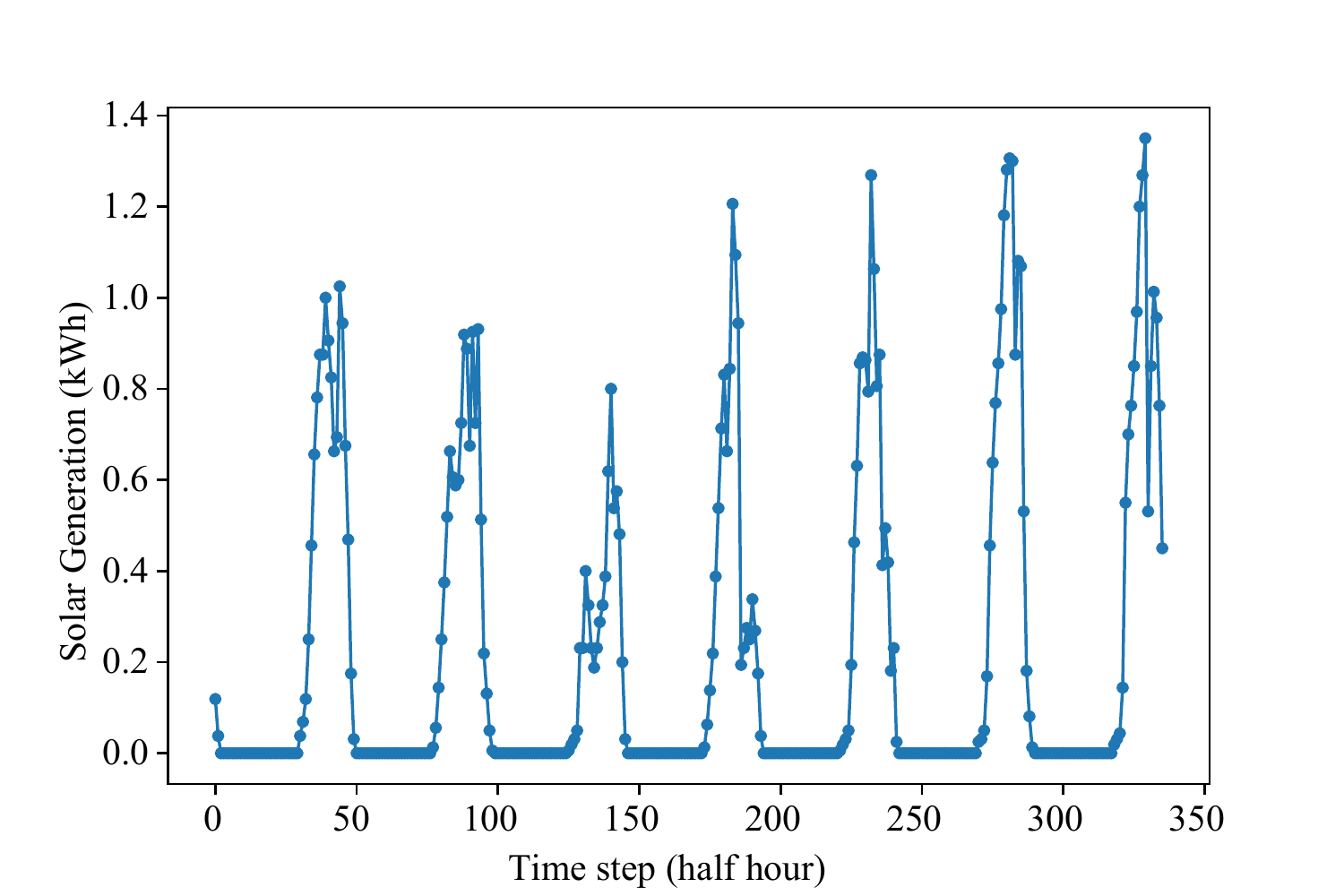}
	\caption{Raw solar generation data used in this paper: first $7$ days}
	\label{fig:dataset}
\end{figure}

The proposed method is implemented using
highly granular solar generation data from Ausgrid distribution network measured using gross meters from 300 individual smart homes having rooftop solar panels \cite{dataset}. The data has been sourced every $30$ minutes in kWh, and thus $48$ solar generation values are available for each day.
In this case study, one year of data from July 1, 2011, to June 30, 2012, is considered for house no. $2076$ and divided into training and testing datasets with $14,052$ and $3,288$ observations, respectively. 
Fig. \ref{fig:dataset} shows the generation behavior for first week ($7$ days) of data to clearly demonstrate the intra-day variability of solar generation. It can be seen in the figure, generation is higher during the peak day hours (noon time), while during night it plunges to $0$ kWh due to unavailability of solar irradiance. 
Furthermore, for peak hours it attains maximum generation upto $1.4$ kWh (day seven), however for day three it attains only $0.8$ kWh reflecting the intermittent nature of renewable energy generation signifying the importance of uncertainty quantification.
\subsection{Experimental setup}
To demonstrate the efficacy of the proposed method, a series of state-of-the-art neural network layers are considered 
with standard Bayesian and VAE architectures.
In specific, M2 (Bayesian bidirectional LSTM), M3 (VAE with Baysian LSTM), M4 (Bayesian LSTM), M5 (VAE with Bayesian RNN), M6 (Bayesian RNN), M7 (VAE with Bayesian ANN), M8 (Quantile regression) are compared against M1 (proposed VAE with Bayesian BiLSTM).
The reason behind this setup is to demonstrate the performance and accuracy of Bayesian probabilistic method with and without the VAE component. 
All the algorithms (M1-M8) are implemented on i7 processor, 16 GB RAM, Nvidia graphics processing unit (GPU) using tensorflow and Keras libraries of python programming language. 
The hyperparameters for the proposed method are tunes using grid search and cross validation as specified in table \ref{hyperparameters}.
Adam \cite{kingma2014adam} is utilized as optimizer to minimize negative log likelihood (NLL) with respect to KL divergence between prior and posterior distribution. While training for $100$ epochs, the technique of early stopping is employed with patience rate of $20$ to prevent the network from over-fitting.
Furthermore, to effectively train the model for unseen data cross validation with $0.2$ split is used monitored with validation loss.
\begin{table}[!t]
	\centering
	\renewcommand{\arraystretch}{1.0}
	\caption{Hyper-parameter settings}\label{hyperparameters}
		\begin{center}
	\begin{tabular}{p{32mm}p{10mm}}	
		\hline
		\hline
		Parameter &Value\\
		\hline
		Optimizer & Adam\\
		Epochs& $100$\\
		Loss & NLL\\
		Learning rate & $0.001$\\
		Batch size & $128$ \\
		Number of neurons&$48$\\
		Activation & tanh \\
		Validation split & $0.2$ \\
		Dropout rate & $0.5$ \\
		Number of historical values & $96$ \\
		Latent dimensions & $48$\\
		\hline
		\hfil
			
	\end{tabular}
\end{center}
\end{table}

\begin{table*}[!t]
	\centering
	\renewcommand{\arraystretch}{1.8}
	\caption{Comparative analysis of proposed method (M1) with one year of solar generation data}\label{table:one}
	Bold values represent best performance in terms of least error
	\begin{tabular}{p{4mm}p{29mm}p{12mm}p{12mm}p{12mm}p{12mm}p{12mm}p{12mm}p{12mm}p{12mm}}
		\hline
		\hline
		Sr. no.&	Method name &RMSE &MAE & R-score& pinball\quad (avg) & Winker score & Brier score \quad & CPU Time (s)&weights\\
		\hline
		
		\textbf{M1}&\textbf{VAE-Bayesian BiLSTM}& $\mathbf{0.0907}$& $\mathbf{0.0450}$&$\mathbf{ 0.9276}$ &$\mathbf{0.1404}$&$\mathbf{0.3855}$ &$\mathbf{0.0235}$&$\mathbf{480.36}$ &$\mathbf{2,022}$\\
		
		{M2} &{Bayesian BiLSTM}  &${0.0998}$&${0.0641}$&${0.9122}$&${0.1424}$&${0.3860}$&${0.0244}$&${600.42}$&${76,422}$\\
		
		M3 &VAE-Bayesian LSTM&$0.0980$&$0.0615$&$0.9154$&$0.1433$&$0.4213$&$0.0257$&$330.75$&$1,062$\\
		
		M4&	Bayesian LSTM \cite{sun2019using} &$0.1053$&$0.0607$&$0.9025$&$0.1821$&$1.0059$&$0.0646$&$230.23$&$38,214$\\
		
		M5 &VAE-Bayesian RNN&$0.1149$&$0.0758$&$0.8838$&$0.1436$&$0.4734$&$0.0329$&$90.69$&$558$\\
		
		M6&Bayesian RNN &$0.3440$&$0.2889$&$-0.039$&$0.2114$&$2.1710$&$0.3653$&$25.30$&$9,990$\\
		M7&Bayesian ANN \cite{yang2019bayesian}&$0.4553$&$0.3990$&$-0.8211$&$0.7902$&$10.643$&$5.8601$&$60.48$&$9,894$\\
		
		M8 & QR &$0.1574$&$0.0727$&$0.7816$&$0.1525$&$1.4447$&$0.0727$&$40.02$&-\\		
		\hline
		\hfill
	\end{tabular}
\end{table*}
\subsection{Evaluation metrics}
To evaluate the proposed VAE-Bayesian BiLSTM and comparative (M2-M8) methods for probabilistic solar generation forecasting, average Pinball loss and Winkler score are computed. To evaluate the approximated distribution,
calibration, reliability, and sharpness are the main factors in the prediction intervals to observe for.
For actual solar generation ($y_{t}$) from the test dataset and predictions ($\hat{y}_{t,q}$) at $t^{th}$ time step , Pinball loss
over percentile $q$ $\in[0,1]$ is formulated as:
	\begin{equation}\label{pinball}
	pinball(y_{t},\hat{y}_{t,q},q)=\begin{cases}
	q (y_{t}-{\hat{y}_{t,q}})  & y_{t} \ge {\hat{y}_{t,q}}\\
	(1-q) ({\hat{y}_{t,q}}-y_{t})  & y_{t} \le {\hat{y}_{t,q}}
	\end{cases}  
	\end{equation}	
	
Furthermore, for confidence $(1-\gamma) \times 100 $, Winkler score is computed using:
\begin{equation}
Winkler=\begin{cases}
\delta & lb_t \le y_t \ge ub_t\\
\delta + 2(lb_t-y_t) / \gamma & lb_t \ge y_t\\
\delta + 2 (y_t-ub_t) / \gamma & ub_t \le y_t\\
\end{cases}
\end{equation}
where $lb_t$ and $ub_t$ represent the lower and upper bounds of probabilistic forecasts at interval $t$, respectively. And, $\delta = ub_t -lb_t $ is the PI width at $t$.
Furthermore, skill score in the terms of Brier Score (BS) \cite{brier1950verification} is also computed to evaluate the error in probabilistic predictions defined as:
\begin{equation}
	BS=\frac{1}{n} \sum_{t=1}^{t=n}(f_t-y_t)^2
\end{equation}
where $f_t$ stands for estimated forecasting values at interval $t$ using approximated posterior distribution and $n$ is the number of samples form testing set.
%
A lower pinball, Winkler, and BS imply better probabilistic estimation.

Furthermore, forecasting accuracy at deterministic level is also evaluated by computing root-mean square error (RMSE) and mean absolute error (MAE) of the differences between actual and predictive mean values (i.e. at $50^{th}$ percentile). The respective equations for RMSE and MAE 
are given as follows:
\begin{equation}
RMSE=\sqrt{\frac{1}{n} \sum\limits_{t=1}^{n}(\hat{y}_{t}-y_{t})^2}
\label{rmse}
\end{equation}
\vspace{-5pt}
\begin{equation}
MAE=\frac{1}{n}\sum\limits_{t=1}^{n}\mid{\hat{y}_{t}-y_{t}}\mid
\label{mape}
\end{equation}
where $\hat{y}_{t}$ and $y_{t}$ symbolize the predictive mean and actual generation at $t$. 
In addition, coefficient of determination symbolized as R is also obtained to represent the goodness of a fit by the forecasting model. It is given as:
\begin{equation}
R= 1- \frac{\sum{(\hat{y}_{t}-y_{t})^2}}{\sum{({y}_{t}-\bar{y})^2}}\label{R}
\end{equation} 
where $\bar{y}$ represents the mean value for actual data points.
\begin{table*}[!ht]
	\centering
	\renewcommand{\arraystretch}{1.8}
	\caption{Performance analysis of proposed method for six months of solar generation data}\label{table:six}
	\begin{tabular}{p{4mm}p{28mm}p{12mm}p{12mm}p{12mm}p{12mm}p{12mm}p{12mm}p{12mm}p{12mm}}
		\hline
		\hline
		Sr. no.&	Method name &RMSE &MAE & R-score& pinball\quad (avg) & Winker score & Brier score \quad  & CPU Time (s)&weights\\
		\hline
		M1&VAE-Bayesian BiLSTM&$ 0.1608$&${0.0994}$&${ 0.8607}$&$0.3345$&$2.6957$&$0.3794$&$144.49$&$2,022$\\
		
		M2&Bayesian BiLSTM&$0.2277$&$0.1925$&$0.7209$&$0.3356$&$3.8757$&$ 0.7316$&$232.81$&$76,422$\\
		& $\%$ Improvement &$29.38\%$&$48.36\%$&$19.39\%$&$0.32\%$&$30.59\%$&$48.14\%$&$37.93\%$&$97.35\%$\\ 
		\hline
		\hfill
	\end{tabular}
\end{table*}
\begin{table*}[!t]
	\centering
	\renewcommand{\arraystretch}{1.8}
	\caption{Performance analysis of proposed method for Intra-Day forecasting (peak 9 hours)}\label{table:intra}
	\begin{tabular}{p{4mm}p{28mm}p{12mm}p{12mm}p{12mm}p{12mm}p{12mm}p{12mm}p{12mm}p{12mm}}
		\hline
		\hline
		Sr. no.&	Method name &RMSE &MAE & R-score& pinball\quad (avg) & Winker score & Brier score& CPU Time (m)&weights\\
		\hline
		M1&VAE-Bayesian BiLSTM&$0.0948$&$0.0741$&$0.9488$&$ 0.4247$&$2.8348$&$0.4750$&$56.92$&$582$\\
		M2&Bayesian BiLSTM&$0.1143$&$0.0943$&$0.9255$&$ 0.5817$&$3.2392$&$0.5294$&$64.37$&$1,494$\\
		&\% Improvement&17.06\%&21.42\%&2.51\%&26.98\%&12.48\%&10.27\%&11.57\%&61.04\%\\
		\hline
		\hfil
	\end{tabular}
\end{table*}
\begin{table*}[!t]
	\centering
	\renewcommand{\arraystretch}{1.8}
	\caption{Comparative analysis of VAE for weight  dimensionality reduction}
	\label{table:recon}
	\begin{tabular}{p{5mm}p{45mm}p{8mm}p{18mm}p{8mm}p{18mm}p{8mm}p{10mm}}
		\hline
		\hline
		\multicolumn{2}{c}{}&\multicolumn{2}{c}{One year} &\multicolumn{2}{c}{Six months}&\multicolumn{2}{c}{Intra-Day}\\
		\hline
		Sr. no. &Method name&Recon. error&Time (s)&Recon. error&Time (s)&Recon. error&Time (s)\\
		\hline
		1 &\textbf{VAE-Bayesian BiLSTM (M1)} &$\mathbf{0.7539}$&$\mathbf{960.49}$&$\mathbf{1.2501}$&$\mathbf{190.52}$&$\mathbf{0.8670}$&$\mathbf{116.28}$\\
		2 &VAE-Bayesian LSTM&$1.0346$&$805.02$&$ 1.7513$&$1223.37$&$1.0400$&$84.26$\\
		3& VAE-Bayesian RNN&$8.515$&$188.36$&$1.8679$&$184.36$&$1.2757$&$32.81$\\
		\hline
	\end{tabular}
\end{table*}
\begin{figure*}[!h]
	\centering
	\subfigure[Predictive mean and PIs for 2 days]
	{\includegraphics[height=0.225\textheight,width=0.5\textwidth]{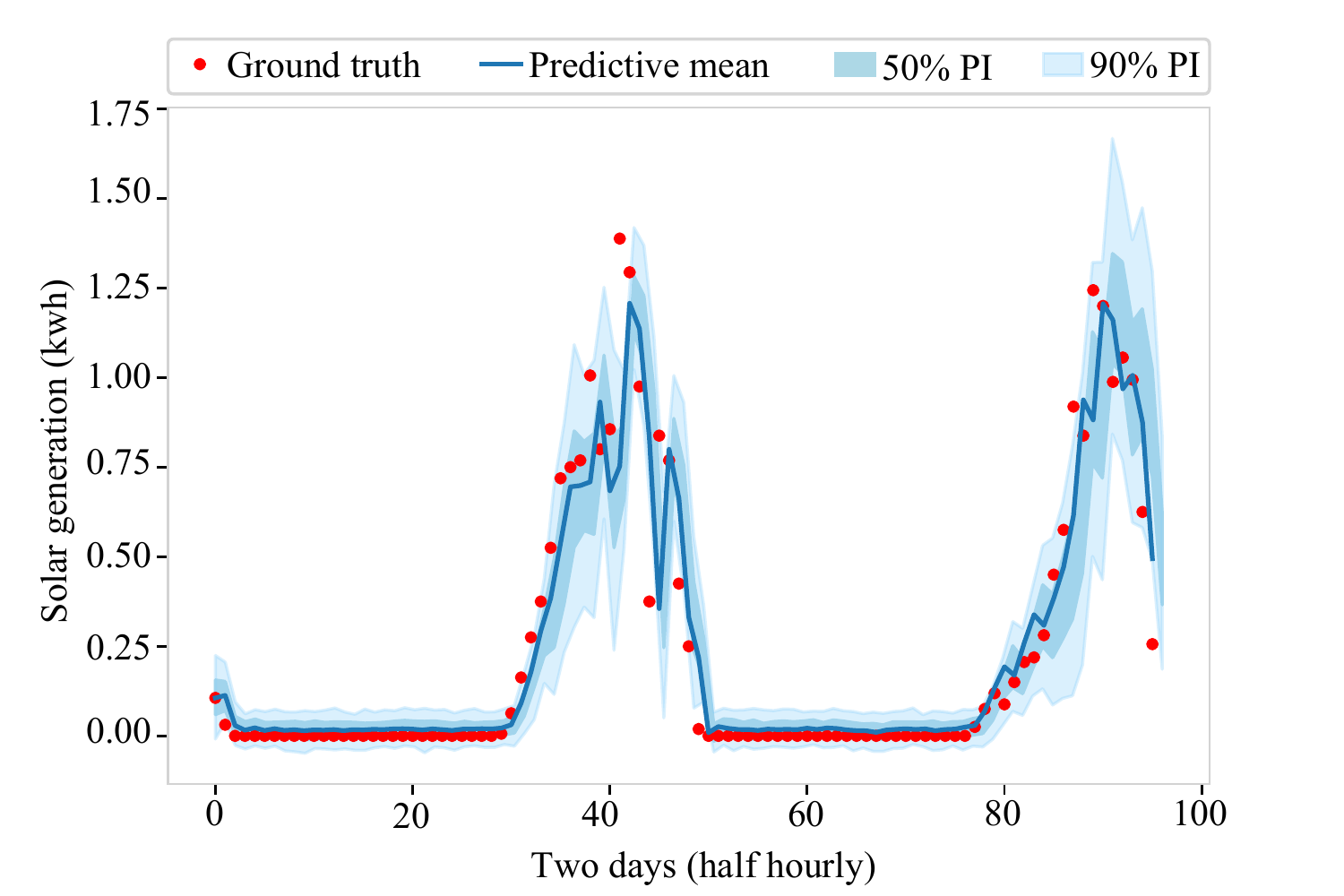}\label{results1a}}
	\hspace{-20pt}
	\subfigure[Predictive mean and PIs for 72 days (entire test periods)]
	{\includegraphics[height=0.225\textheight,width=0.5\textwidth]{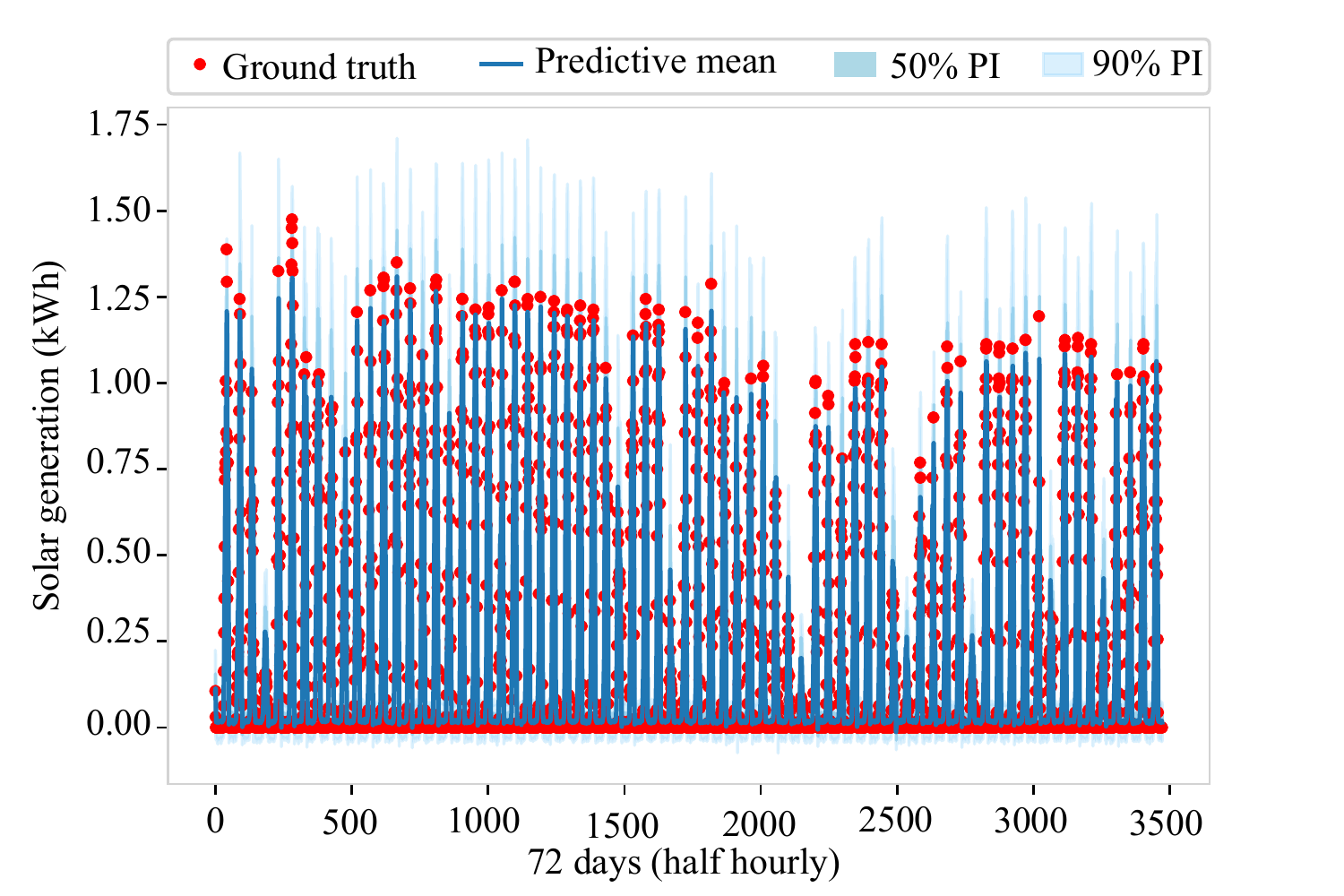}\label{results1b}}
	\caption{Forecasting results for proposed VAE-Bayesian BiLSTM (M1)}
	\label{results1}
\end{figure*}
\begin{figure*}[!h]
	\centering
	\subfigure[Predictive mean and PIs for 2 days]	
	{\includegraphics[height=0.225\textheight,width=0.5\textwidth]{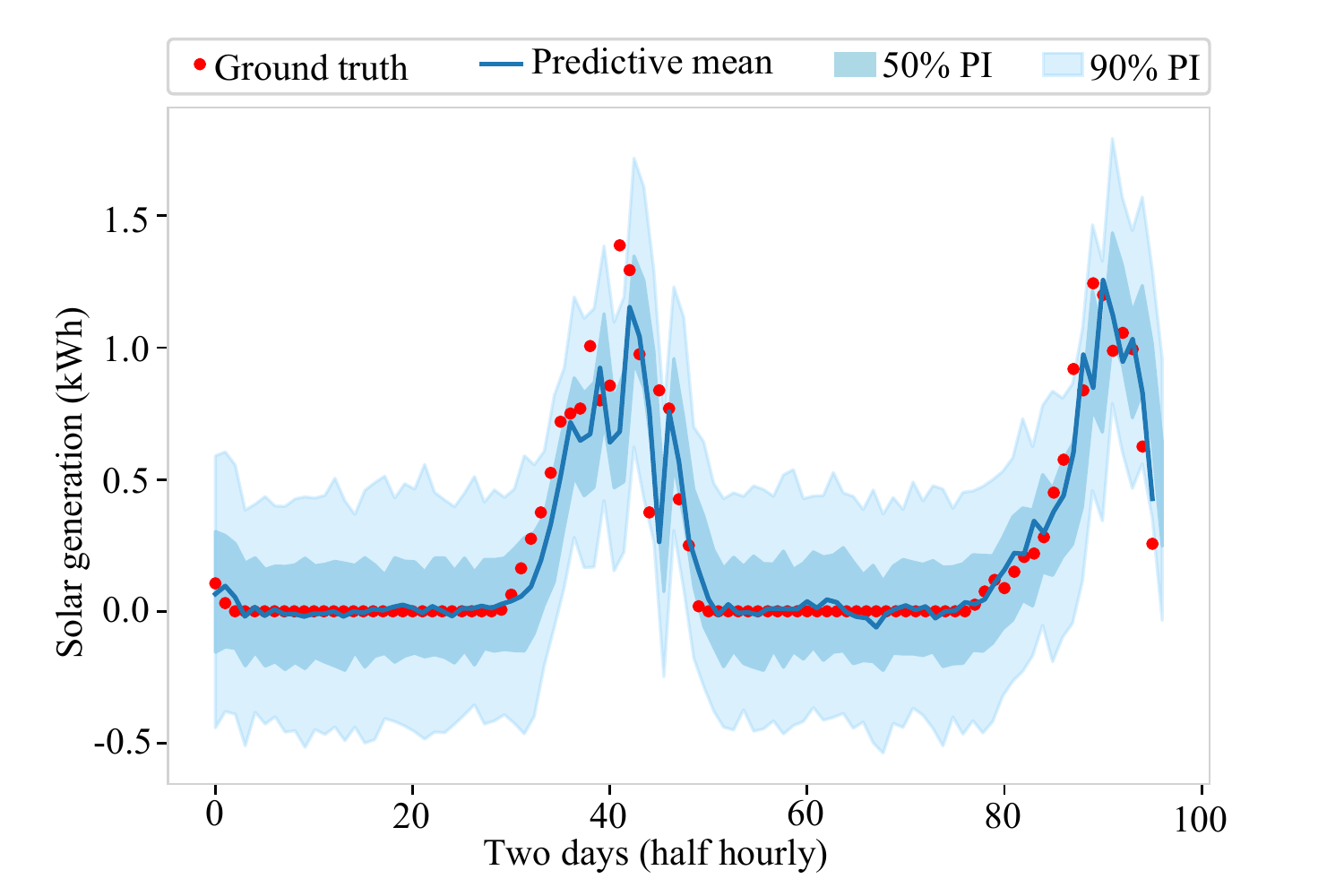}}
	\hspace{-20pt}	
	\subfigure[Predictive mean and PIs for 72 days (entire test periods)]
	{\includegraphics[height=0.225\textheight,width=0.5\textwidth]{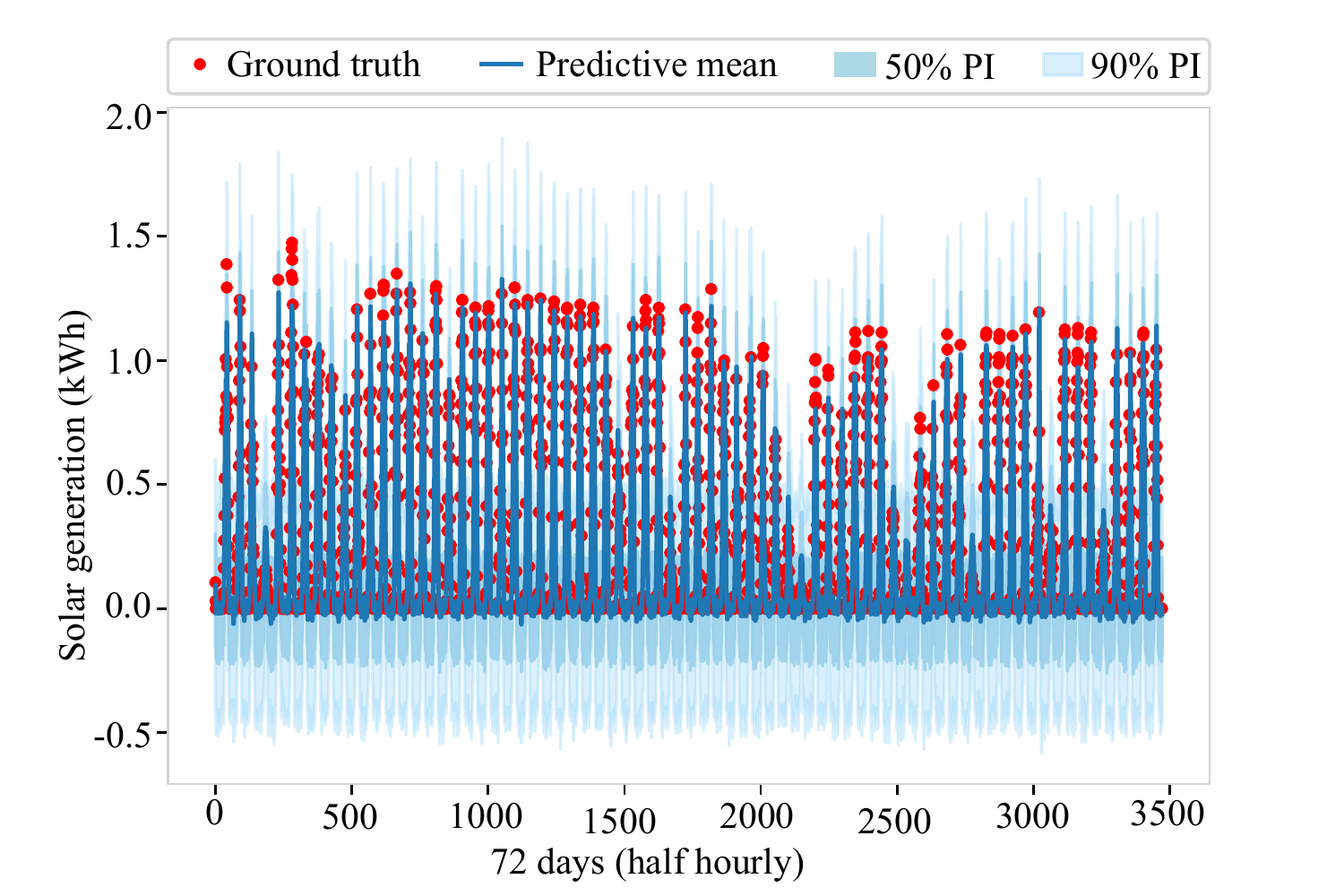}}
	\caption{Forecasting results for Bayesian BiLSTM (M2)}
	\label{results2}
\end{figure*}

\subsection{Framework Evaluation and Comparative Analysis}
In this subsection, we present the implementation results for the proposed VAE-Bayesian BiLSTM framework.
An extensive comparative analysis is conducted to evaluate the performance of proposed method with state-of-the-art deep learning-based probabilistic algorithms (M2-M7) as reflected in Table \ref{table:one}. 
It can be observed from the table that the least error values are reflected by the proposed VAE-Bayesian BiLSTM in the terms of RMSE, MAE, pinball, Winkler, and BS.
Also, Bayesian BiLSTM, LSTM, and RNN layers exhibit a reduction in aforementioned prediction errors when combined with VAE.
It is important to note that the number of weight parameters significantly decrease for all the Bayesian algorithms when integrated with the VAE component. In this regard, M1 which is our proposed algorithm, reflects a considerable weight reduction from $76,4224$ (when VAE is not integrated) to $2,022$ (with VAE) to train the forecasting framework. 

As reflected in the table \ref{table:one}, Proposed method (M1) outperforms VAE-Bayesian LSTM and VAE-Bayesian RNN in terms of forecasting error and pinball loss highlighting its best performance for probabilistic forecasting and uncertainty quantification.
Additionally, higher values of R-score for BiLSTM-based methods justify the goodness of fit and learning performance achieved using bidirectional LSTM layers.
Note that the reported CPU execution time in the results refers to the learning process explicitly. Execution time for common computations between all methods is not included such as for data normalization, error computation, etc. Also, the best results are summarized in this section after appropriate hyperparameter tuning, for brevity.
Furthermore, we affirm the efficacy of our proposed method by performing an analysis with different data sizes in comparison to the Bayesian BiLSTM method. In this regard, tables \ref{table:six} and \ref{table:intra} compare above two methods on the basis of evaluation metrics for six months and intra-day solar generation data, respectively. 
It can be observed that integration with VAE significantly improves the weight dimensionality and forecasting accuracy for both sizes of datasets though the improvement in pinball score is more for intra-day forecasting. This clearly indicates that the proposed method can also achieve superior performance for intra-day forecasting.

Note that a trade-off is observed between the two methods for time complexity and forecasting error, as VAE contributes to some information loss during data encoding.
In this regard, it is constructive to outline that while employing VAE, reconstruction error has to be monitored and minimized with respect to the error threshold value. The reconstruction error and execution time for the VAE component with BiLSTM, LSTM and RNN are demonstrated in Table \ref{table:recon}. Although, BiLSTM takes slightly more time for training as compared to LSTM and RNN due to bidirectional processing, the values clearly reflect that the proposed method achieve minimum error for data reconstruction.

Furthermore, Fig. \ref{results1a} and Fig. \ref{results1b} demonstrate two days and $72$ days of forecasting horizons respectively, plotted with ground truth from test set and predictive mean given by proposed VAE-Bayesian BiLSTM method. Additionally, $90 \%$ and $50 \%$ PIs are plotted which reflect the future probabilities on different percentiles in relation to the predictive mean and ground truth. It is inferred from the figures that the proposed method is capable to quantify uncertainties through tighter bounds on future probabilities in the form of PIs.
Similarly, Fig. \ref{results2} exhibits the PIs generated by Bayesian BiLSTM on aforementioned two time horizons. From the graphs, it is clear that state-of-the-art Bayesian algorithm with BiLSTM has wider prediction coverage for future probabilities corresponding to less reliability and sharpness in comparison to the proposed method.
Based on above results, it can be concluded that the proposed VAE-Bayeisan BiLSTM method effectively quantifies the uncertainties with tighter PIs and lower computational cost. However, a trade-off needs to be maintained between the computational efficiency and forecasting performance, besides maintaining a lower reconstruction error.
In addition, table \ref{table:ratiochanges} presents tain-test ratio variations for the proposed method. As we increase the training size the prediction errors may decrease, however it will contribute to the problem of overfitting.
\begin{table}[!h]
	\centering
	\renewcommand{\arraystretch}{1.8}
	\caption{Impact of different train-test variations on prediction errors}\label{table:ratiochanges}
	\begin{tabular}{p{5mm}p{8mm}p{8mm}p{8mm}p{8mm}p{8mm}p{8mm}}
		\hline
		\hline
		Ratio &RMSE &MAE & R-score& Pinball\quad (avg) & Winker score & Brier score \\
		\hline
		80:20&${0.0907}$& ${0.0450}$&${0.9276}$ &${0.1404}$&${0.3855}$ &${0.0235}$\\
		
		70:30&$0.0902$&$0.0425$&$0.9376$&$0.1400$&$0.3825$&$0.0233$\\
		
		60:40&$0.0889$&$0.0315$&$0.9566$&$0.1325$&$0.2245$&$0.0143$\\
		50:50 &$0.0876$&$0.0302$&$0.9573$&$0.1315$&$0.2125$&$0.0132$\\
		
		\hline
		\hfill
	\end{tabular}
\end{table}
\section{Conclusion}\label{sec:conclusions}
In this paper, a new VAE-based Bayesian BiLSTM technique for renewable energy generation forecasting is presented to quantify the model and stochastic uncertainties, while optimising the number of weight parameters to improve the computational efficiency. The proposed technique outperforms benchmark point and PDL forecasting techniques in terms of RMSE and pinball loss. Numerical results presented in the results section demonstrate the superior forecasting and computational performance of the proposed method in comparison to the other PDL methods. Furthermore, uncertainties are addressed by proposed method efficiently in the form of future prediction intervals.
\bibliographystyle{cas-model2-names}

\end{document}